\documentclass[11pt,a4paper]{article}
\usepackage[hyperref]{acl2017}
\usepackage{times}
\usepackage{latexsym}
\usepackage[english]{babel}
\usepackage{blindtext}

\usepackage{url}

\aclfinalcopy 

\title{Who's to say what's funny?\\
A computer using Language Models and Deep Learning, That's Who!}

\author{Xinru Yan \& Ted Pedersen\\
  Department of Computer Science \\ 
  University of Minnesota Duluth \\ 
   Duluth, MN, 55812 USA \\
  {\tt \{yanxx418,tpederse\}@d.umn.edu}}

\date{}

\begin{document}
\maketitle
\begin{abstract}
Humor is a defining characteristic of human beings. 
Our goal is to develop methods that automatically 
detect humorous statements and rank them on a 
continuous scale. In this paper we report on results 
using a Language Model approach, and outline our
plans for using methods from Deep Learning.
\end{abstract}

\section{Introduction}
\textit{Computational humor} is an emerging area of 
research that ties together ideas from psychology, 
linguistics, and cognitive science. 
\textit{Humor generation} is the problem of automatically
creating humorous statements 
(e.g., \cite{StockS03}, \cite{OzbalS12}).
\textit{Humor detection} seeks to identify humor in text, 
and is sometimes cast as a binary classification problem that 
decides if some input is humorous or not 
(e.g., \cite{MihalceaS06}, 
\cite{ZhangL14}, \cite{ShahafHM15},
\cite{MillerG15}). However, our focus is on  
the continuous and subjective aspects of humor. 

We learn a particular sense of humor from a data set of
tweets which are geared towards a certain style of humor
\cite{PotashRR16}. 
This data consists of humorous tweets which have been
submitted in response to hashtag prompts provided
during the Comedy
Central TV show \textit{@midnight with Chris Hardwick}.
Since not all jokes are equally funny, we use Language
Models and methods from Deep Learning to allow 
potentially humorous statements to be ranked relative to 
each other. 

\section{Language Models}

We used traditional Ngram language models as our first 
approach for two reasons : 
First, Ngram language models can learn a certain style of 
humor by using examples of that as the training data for the model.
Second, they assign a probability to each input they are given,
making it possible to rank statements relative to each other.  
Thus, Ngram language models make relative rankings 
of humorous statements
based on a particular style of humor, thereby accounting for the 
continuous and subjective nature of humor. 

We began this research by participating in SemEval-2017 Task 6 
\#HashtagWars: Learning a Sense of Humor \cite{PotashRR17}. This included
two subtasks : Pairwise Comparison (Subtask A) and Semi-ranking 
(Subtask B). Pairwise comparison asks a system to choose the funnier of
two tweets. Semi-ranking requires that each of the tweets
associated with a particular hashtag be assigned to one of the 
following categories : top most funny tweet, next nine most funny
tweets, and all remaining tweets. 


Our system
estimated tweet probabilities using Ngram language models.
We created models
from two different corpora - a collection of funny tweets from the @midnight
program, and a corpus of news data that is freely available for 
research\footnote{http://www.statmt.org/wmt11/featured-translation-task.html}. 
We scored tweets by assigning them a probability based on each 
model. Tweets that have a higher probability according to the funny 
tweet model are considered funnier since they are more like the humorous
training data. However, tweets 
that have a lower probability according to the news language model 
are viewed as funnier since they are least like the (unfunny) news corpus.
We took a standard approach to language modeling and used bigrams 
and trigrams as features in our models. We used KenLM \cite{HeafieldPCK13} 
with modified Kneser-Ney smoothing and a back-off technique as our language 
modeling tool. 

Table 1 shows our results for both data sets when trained on 
bigrams and trigrams. The accuracy and distance measures
are defined by the task organizers \cite{PotashRR17}. We seek 
high accuracy in picking the funnier tweet (Subtask A) and low
distance (from the gold standard) in organizing the tweets into 
categories (Subtask B). 

\begin{table}[h]
\begin{center}
\begin{tabular}{cccc}
\hline
Data & Ngram & Accuracy (A) & Distance (B) \\
\hline
tweets & trigram & 0.397 & 0.967 \\
tweets & bigram & 0.406 & 0.944 \\
news & trigram & 0.627 & 0.872 \\
news & bigram & 0.624 & 0.853 \\
\end{tabular}
\caption{Experimental results}
\end{center}
\end{table}

These results show that models trained on the news data
have a significant advantage over the tweets model, and that bigram
models performed slightly better than trigrams. We submitted
trigram models trained on news and tweets to the official evaluation 
of SemEval-2017 Task 6. The trigram language models trained on
the news data placed fourth in Subtask A and first in Subtask B.

We believe that the significant advantage of the news data over 
the tweet data is caused by the much larger quantity of news data
available. The tweet data only consists of approximately 21,000 tweets,
whereas the news data totals approximately 6.2 GB of text.
In the future we intend to collect more tweet data, especially those 
participating in the ongoing \#HashtagWars staged nightly by @midnight. 
We also plan to experiment with equal amounts of tweet data and
news data, to see if one has an inherent advantage over the other.

Our language models performed better in the pairwise comparison, 
but it is clear that more investigation is needed to improve
the semi-ranking results. 
We believe that Deep Learning
may overcome some of the limits of Ngram language models, and so
will explore those next. 

\section{Deep Learning}

One limitation of our language model approach  
is the large number of out of vocabulary 
words we  encounter. This problem can not be solved
by increasing the quantity of training data because
humor relies on creative use of language. 
For example, jokes often include puns based on 
invented words, e.g., a singing cat makes beautiful {\textit meowsic}.  
\cite{PotashRR16} suggests that character--based 
Convolutional Neural Networks (CNNs) are
an effective solution for these situations since they are not
dependent on observing tokens in training data.
Previous work
has also shown the CNNs are effective tools for
language modeling, even in the presence of complex morphology 
\cite{KimJSR15}. Other recent work has shown that Recurrent
Neural Networks (RNNs), in particular Long Short--Term Memory
networks (LSTMs), are effective in a wide range of language
modeling tasks (e.g., \cite{SundermeyerSN12},\cite{SundermeyeNS15}). 
This seems to be due to their ability to capture
long distance dependencies, which is something that Ngram language
models can not do. 

\cite{PotashRR16} finds that 
external knowledge is necessary to detect humor in tweet
based data. This might include information about book and movie
titles, song lyrics, biographies of celebrities etc. and is
necessary given the reliance on current events and
popular culture in making certain kinds of jokes.

We believe that Deep Learning techniques potentially offer
improved handling of unknown words, long distance dependencies
in text, and non--linear relationships among words and
concepts. Moving forward we intend to explore a variety of
these ideas and describe those briefly below.

\section{Future Work}

Our current language model approach is effective but 
does not account for out of vocabulary
words nor long distance dependencies. CNNs in combination with
LSTMs seem to be a particularly promising way to overcome 
these limitations (e.g., \cite{BerteroF16}) which we will
explore and compare to our existing results.

After evaluating
CNNs and LSTMs we will explore how to include domain knowledge
in these models. One possibility is to
create word embeddings
from domain specific materials and provide those to the CNNs
along with more general text. Another is to investigate the use
of Tree--Structured LSTMs \cite{TaiSM15}. These have the potential
advantage of preserving non-linear structure in text, which may
be helpful in recognizing some of the unusual variations of words
and concepts that are characteristic of humor.


\end{document}